\documentclass[runningheads]{llncs}
\usepackage{graphicx}
\usepackage{tikz}
\usepackage{comment}
\usepackage{amsmath,amssymb} % define this before the line numbering.
\usepackage{color}
\usepackage{enumerate}
\usepackage{algorithm, algorithmic}
\usepackage{multirow}
\usepackage[normalem]{ulem}
\useunder{\uline}{\ul}{}
\usepackage{arydshln}

\usepackage[accsupp]{axessibility}  % Improves PDF readability for those with disabilities.

\begin{document}
% \linenumbers
\pagestyle{headings}
\mainmatter
\def\ECCVSubNumber{3041}  % Insert your submission number here

\title{Translation, Scale and Rotation: Cross-Modal Alignment Meets RGB-Infrared Vehicle Detection}  % Replace with your title

\begin{comment}
\titlerunning{ECCV-22 submission ID \ECCVSubNumber} 
\authorrunning{ECCV-22 submission ID \ECCVSubNumber} 
\author{Anonymous ECCV submission}
\institute{Paper ID \ECCVSubNumber}
\end{comment}
%******************

%\begin{comment}
\titlerunning{Cross-Modal Alignment Meets RGB-Infrared Vehicle Detection}
\author{Maoxun Yuan\inst{1} \and
Yinyan Wang\inst{1} \and
Xingxing Wei\inst{2}\thanks{Corresponding author.}}
\authorrunning{Maoxun Yuan et al.}
\institute{Beijing Key Laboratory of Digital Media, Beihang University, Beijing, China \and
Institute of Artificial Intelligence, Hangzhou Innovation Institute, Beihang University, Beijing, China\\
\email{\{yuanmaoxun,wangyinyan,xxwei\}@buaa.edu.cn}}
%\end{comment}
%******************

\maketitle

\begin{abstract}
Integrating multispectral data in object detection, especially visible and infrared images, has received great attention in recent years. Since visible (RGB) and infrared (IR) images can provide complementary information to handle light variations, the paired images are used in many fields, such as multispectral pedestrian detection, RGB-IR crowd counting and RGB-IR salient object detection. Compared with natural RGB-IR images, we find detection in aerial RGB-IR images suffers from cross-modal weakly misalignment problems, which are manifested in the position, size and angle deviations of the same object. In this paper, we mainly address the challenge of cross-modal weakly misalignment in aerial RGB-IR images. Specifically, we firstly explain and analyze the cause of the weakly misalignment problem. Then, we propose a Translation-Scale-Rotation Alignment (TSRA) module to address the problem by calibrating the feature maps from these two modalities. The module predicts the deviation between two modality objects through an alignment process and utilizes Modality-Selection (MS) strategy to improve the performance of alignment. Finally, a two-stream feature alignment detector (TSFADet) based on the TSRA module is constructed for RGB-IR object detection in aerial images. With comprehensive experiments on the public DroneVehicle datasets, we verify that our method reduces the effect of the cross-modal misalignment and achieve robust detection results.
\end{abstract}

\section{Introduction}
Object detection in aerial images plays an important role in computer vision field with various applications, such as urban planning, surveillance and disaster rescue. Unlike natural images that are often taken from low-altitude perspectives, aerial images are typically taken with bird views, which implies that objects in aerial images are always distributed with arbitrary orientation. To solve these problems, several oriented object detectors \cite{ding2019learning,wang2019mask,xie2021oriented,zhou2020arbitrary} have been proposed and obtained state-of-the-art results on challenging aerial image datasets  \cite{liu2017high,xia2018dota}. However, these detectors are only designed for the visible images alone, which cannot cope with the challenges in limited illumination (nighttime).

For these reasons, infrared cameras have been invested to deal with complex scenarios. Infrared cameras can present clear silhouettes of objects even in low-light condition, due to the capability in capturing the radiated heat. This makes visible (RGB) images and infrared (IR) images complement each other. In natural images, RGB-IR images are utilized in many fields , such as multispectral pedestrian detection, RGB-IR crowd counting  and RGB-IR salient object detection. However, there are few RGB-IR methods and paired image datasets specifically for aerial imagery. 
% However, compared with natural images, there are few RGB-IR paired datasets and RGB-IR fusion schemes for aerial images. Recently, Sun et al. \cite{sun2020drone} first collect a large-scale drone-based RGB-IR dataset (\emph{DroneVehicle}), which contains low-light and adverse weather (fog, rain, dust) conditions. The dataset has been performed the registration algorithm to provide aligned image pairs. The release of DroneVehicle dataset paves the way to promote the development of cross-modal object detection in aerial imagery.

\begin{figure*}[!t]
	\begin{center}
	\includegraphics[scale=0.26]{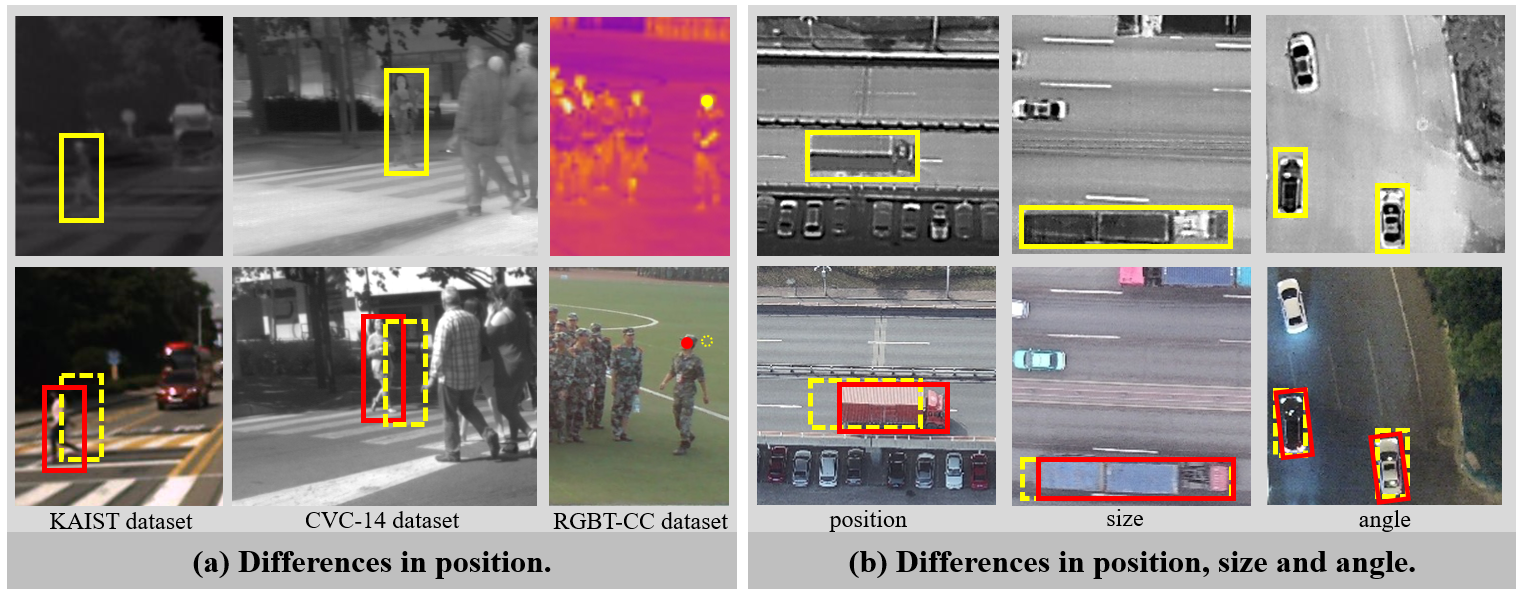}
	\end{center}
    \caption{Illustration of the modality weakly misalignment problem. (a) and (b) are the visualization image patches (cropped on the same position of RGB-IR image pairs) of groundtruth annotations in nature image datasets and aerial DroneVehicle dataset. The yellow and the red boxes represent annotations of same objects in the infrared images and the visible images, respectively. KAIST and CVC-14 are the pedestrian detection dataset, and RGBT-CC is the crowd counting dataset.}
	\label{fig1}
\end{figure*}

Image alignment is one of the issues that should be considered in cross-modal image applications. Existing methods \cite{li2019rgb,liu2021cross,zhou2020improving,xu2017learning,guan2019fusion,zhang2019rgb,zhang2021abmdrnet} usually assume that visible-infrared image pairs are perfectly geometrically aligned, and they directly perform the multi-modal fusion methods. Actually, after the image registration algorithm, the paired images are just weakly aligned (shown in Fig.~\ref{fig1}(a)). However, objects are always arbitrary oriented in aerial images, they differ not only in position, but also in scale and angle. These three deviations (position, size and angle) of the paired arbitrary oriented objects are closely coupled, changing one will affect another one, which makes the alignment operation more complicated. As shown in Fig.~\ref{fig1}(b), the dashed boxes represent the position of the corresponding bounding-boxes in the infrared images. We find that the same object on image pairs differs in location, scale and angle. Therefore, weakly misalignment in cross-modal aerial images is a common issue that needs to be addressed.
% In addition, we separately count the number of bounding-boxes with deviations (position and size offset by 3 pixels, angle offset by 3 degrees) in the DroneVehicle dataset. As illustrated in Fig.~\ref{fig1}(b), more than 20\% of the bounding-boxes have the deviation problem. Therefore, weakly misalignment in cross-modal aerial images is a common issue that needs to be addressed.

To solve the above problem, in this paper, we propose and analyze the cause of the weakly misalignment problem. The problem is mainly caused by two factors: hardware errors and annotation errors. To address these issues, we propose a TSRA module to calibrate the feature maps from two modality proposals through translation, scale and rotation operations. In the module, the feature maps of RGB-IR modality proposals are first subjected to the alignment process to acquire the deviation, and then we utilize a Modality-Selection strategy to select the appropriate annotations as the reference bounding-boxes. The final feature maps for classification and regression are obtained by fusing the aligned proposal features. Finally, we construct a TSRA-based oriented object detector to evaluate the effectiveness of the TSRA module.
% Finally, we construct a framework based on TSRA module to cope with the difficulty of weakly alignment in RGB-IR object detection.

In summary, the contributions of this paper are as follows:
\begin{itemize}
    \item[-] We present the cross-modal weakly misalignment problem specific to the RGB-IR object detection in aerial images. To the best of our knowledge, it is the ﬁrst time to present and analyse the weakly misalignment problem in rotated object detection of RGB-IR aerial images. 
    \item[-] We propose a TSRA module which consists of alignment process and MS strategy to translation, scale and rotate the feature maps of two modality objects. Meanwhile,  the Multi-task Jitter is designed to further improve model performance. To evaluate the validity of the TSRA module, we construct a two-stream feature alignment detector  (TSFADet) for RGB-IR object detection in aerial images, which can be trained with an end-to-end manner.
    \item[-] Extensive experiments on DroneVehicle dataset demonstrate that, our TSFADet outperforms previous state-of-the-art datectors and the TSRA module is effective for solving the cross-modal weakly misalignment problem.
\end{itemize}

% The rest of the paper is organized as follows. In the next section we review the scientific literature related to our proposed approach. Section \ref{Method} analyses the problem and describes our approach. We report in section \ref{Experiment} on an extensive set of experiments performed to evaluate the effectiveness of TSFADet, and in section  \ref{Conclusions} we conclude with a discussion of our contribution.

\section{Related Work}
\subsection{Oriented Object Detection}
Aerial images are the main application scenarios of the rotation detectors. Xia et al.  \cite{xia2018dota} construct a large-scale object detection benchmark with oriented annotations, named DOTA. Since then, several existing works  \cite{liu2016ship,ma2018arbitrary,ding2019learning,li2019feature,xu2020gliding,xie2021oriented} are mainly based on typical proposal-based frameworks to explore oriented object detection. Naturally, some methods \cite{liu2016ship,ma2018arbitrary} set numerous rotated anchors with different angles, scales and aspect ratios for better regression. These methods lead to extensive computation complexity. To avoid a large number of anchors, Ding et al. \cite{ding2019learning} designed an RoI transformer to learn the transformation from Horizontal RoIs (HRoIs) to Rotated RoI (RRoIs), which boosts the detection accuracy of oriented objects. 
% To address the challenges of small, cluttered, and rotated object detection, Yang et.al \cite{yang2019scrdet} presented an oriented object detector with the generic object detection framework of Faster R-CNN. 
Recently, Oriented R-CNN \cite {xie2021oriented} is proposed to further improve the detection performance by replacing RROI learning module with a lighter and simpler oriented region proposal network (orientation RPN).  

To improve real-time and availability of detectors, some works \cite{han2021align,yang2019r3det,pan2020dynamic,wei2020oriented,yi2021oriented} have explored one-stage or anchor-free oriented object detection frameworks. For instance, $\rm R^3$Det \cite{yang2019r3det} and $\rm S^2$A-Net \cite{han2021align} are one-stage object detector, which align the feature between horizontal receptive fields and rotated anchors. Recently, GWD \cite{yang2021rethinking} and KLD  \cite{yang2021learning} are proposed to use the Gaussian Wasserstein distance and KL divergence to optimize the localization of the bounding boxes, respectively. 
% Some CenterNet-based methods \cite{pan2020dynamic,wei2020oriented} show their advantages in detecting small objects. As an anchor-free oriented object detector, DRN is devised with a dynamic feature selection and refinement network.
Considering that many current mainstream rotation detectors are based on the well-extended two-stage detection framework, we also build a two-stream rotation detector based on the two-stage detection framework to verify the effectiveness of our proposed method.
% Based on these mainstream detection methods, in this paper, we build a two-stream rotated detector to verify the effectiveness of our proposed method.
% Considering that many current mainstream rotation detectors are based on the well-extended two-stage detection framework, we also build a two-stream rotation detector based on the two-stage detection framework to verify the effectiveness of our proposed method.

\begin{figure*}[!t]
	\begin{center}
	\includegraphics[scale=0.36]{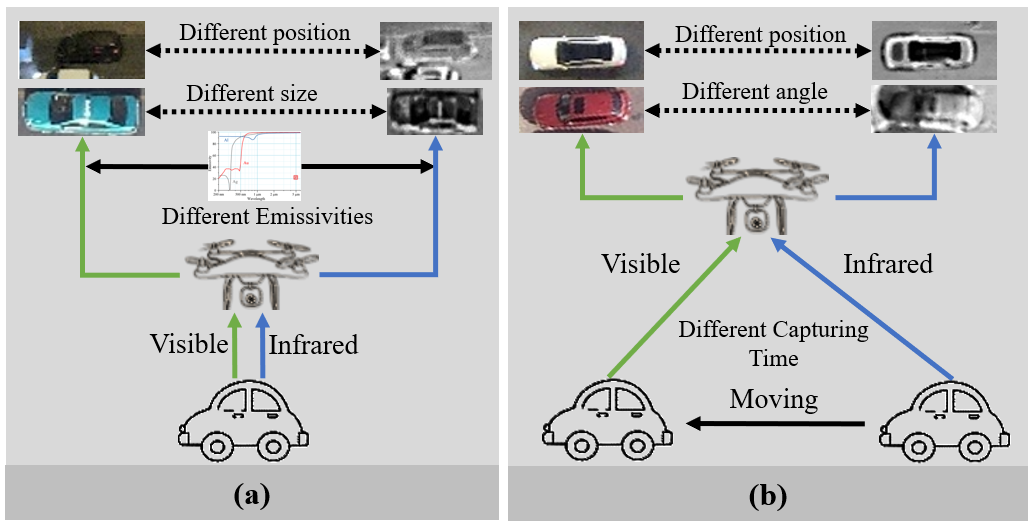}
	\end{center}
    \caption{The illustration of Hardware Errors. (a) and (b) are the description of radiation distortion and clock skew, respectively.}
	\label{fig2_a}
\end{figure*}

\subsection{Cross-Modal Image Alignment}
Images from different modalities usually contain scale, rotation or radiance differences so that cross-modal image alignment is required before using them simultaneously \cite{brown1992survey,zitova2003image}. The aim of image alignment is to warp a sensed image into the common spatial coordinate system of a reference image so that they are matched in pixel. Existing methods are generally divided into area-based methods and feature-based methods. The area-based methods register the image pairs using a similarity metric function, while the feature-based methods include four processes: feature extraction, feature matching, transformation model estimation and image re-sampling and warping. As deep learning has great potential in feature extraction, numerous researchers have designed data-driven strategy in the field of cross-modal image alignment  \cite{zhang2021explore,ma2019novel,cui2021map}. Although image alignment is a necessary step in many fields, it brings extra time consumption and cannot completely address the weakly misalignment problem.
% Although image alignment is a necessary step in many fields, it brings extra time consumption and break the original end-to-end process when used as a preprocessing method. More importantly, we can also see from the DroneVehicle dataset that these traditional methods cannot completely address the weakly misalignment problem.

Recently, a few works have been proposed to address the image alignment issue by end-to-end training network. \cite{zhang2019weakly} firstly addressed the alignment problem by introducing region feature alignment (RFA) module. Zhou et al. \cite{zhou2020improving} designed a illumination aware feature alignment (IAFA) module to align two modality features and then construct Modality Balance Network (MBNet) based on SSD. However, these methods only consider the translation way to solve the deviation, and cannot solve complex misalignment problems in aerial images, such as angle deviation and size deviation. Inspired by \cite{zhang2019weakly}, in this paper, we propose TSRA module to predict the offset of two objects in position, size and angle.

\section{Methodology and Analysis} \label{Method}
In this section, we first analyze the cause of modality weakly misalignment problem (Section \ref{analysis}). According to our analysis, we propose the TSRA module (Section  \ref{tsra-module}) which consists of alignment process, MS strategy and multi-task jitter  to solve the problem. For the same object in two modality images, we use MS strategy to select the bounding-boxes with better annotation in the two modalities as the reference modality, and then perform alignment process to calibrate the feature maps from two modalities. Finally, we put everything together into a description of the TSFADet (Section \ref{overall-architecture}).

\subsection{Analysis} \label{analysis}
Modality weakly misalignment is a common problem in aerial cross-modal images, since the data are collected by different sensors. Through observation and research, we find the the problem usually occurs in the three situation: hardware errors, annotation errors and both, which are explained as follows. 
% The cause of the problem can be summarized into three factors: radiation distortions, clock skews and annotation errors.

\subsubsection{Hardware Errors:}
Hardware errors are mainly reflected in radiation distortions and clock skews.
The radiation distortions often occur in the process of the sensor imaging  \cite{richards1999remote}. The spectral emissivity of the ground objects is different from the real spectral emissivity. These radiation differences will cause images have different representations (\emph{e.g.} color, intensity and texture) for the same objects (shown in Fig.~\ref{fig2_a}(a)). Therefore, the same object on two modality images will have differences in scale and position caused by radiation distortion (shown in  Fig.~\ref{fig1}(a)).

As shown in Fig.~\ref{fig2_a}(b), due to the different sensors' capturing time, the clock skew between two sensors (\emph{e.g.} visible and infrared) can lead to pixel-misalignment of image pairs, especially for locally moving objects such as cars on a highway  \cite{lee2021sipsa}. As a result, clock skew causes the position and angle of the same object to be inconsistent in different modalities, see Fig.~\ref{fig1}(a). 

\begin{figure*}[!t]
	\begin{center}
	\includegraphics[scale=0.40]{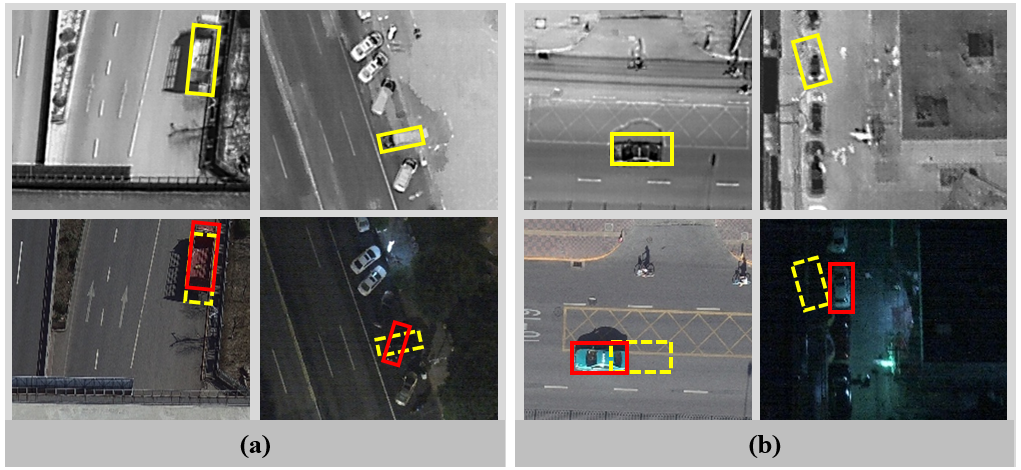}
	\end{center}
    \caption{The visualization image patches (cropped on the same position of RGB-IR image pairs) in the DroneVehicle dataset. (a) Examples of annotation errors. (b) Examples of hardware errors and annotation errors occur simultaneously.}
	\label{fig2_b}
\end{figure*}

\subsubsection{Annotation Errors:}
Since different people have different standards for labeling data, the annotation errors are inevitable. In the process of multispectral data annotation, it is difficult to ensure that the objects are annotated accurately in different modalities. The annotations can affect the training performance of the model. Some examples of the annotation errors in DroneVehicle datasat are shown in Fig.~\ref{fig2_b}(a).

Hardware errors and annotation errors can occur simultaneously in the same object, as shown in Fig.~\ref{fig2_b}(b). That makes the deviation between two modalities cannot be simply solved by affine transformation, therefore, we need to  design a module to handle the above situations and perform alignment process in a  region-wise way.

\subsection{Translation-Scale-Rotation Alignment Module} \label{tsra-module}
The proposed TSRA module can be injected into the object detection framework to solve the weakly misalignment problem. The module mainly consists of two parts: alignment process and modality-selection strategy. To improve the robustness of the TSRA module, we also present a multi-task jitter to augment the alignment process. 

\subsubsection{Alignment Process.} 
\begin{figure*}[!t]
	\begin{center}
	\includegraphics[scale=0.24]{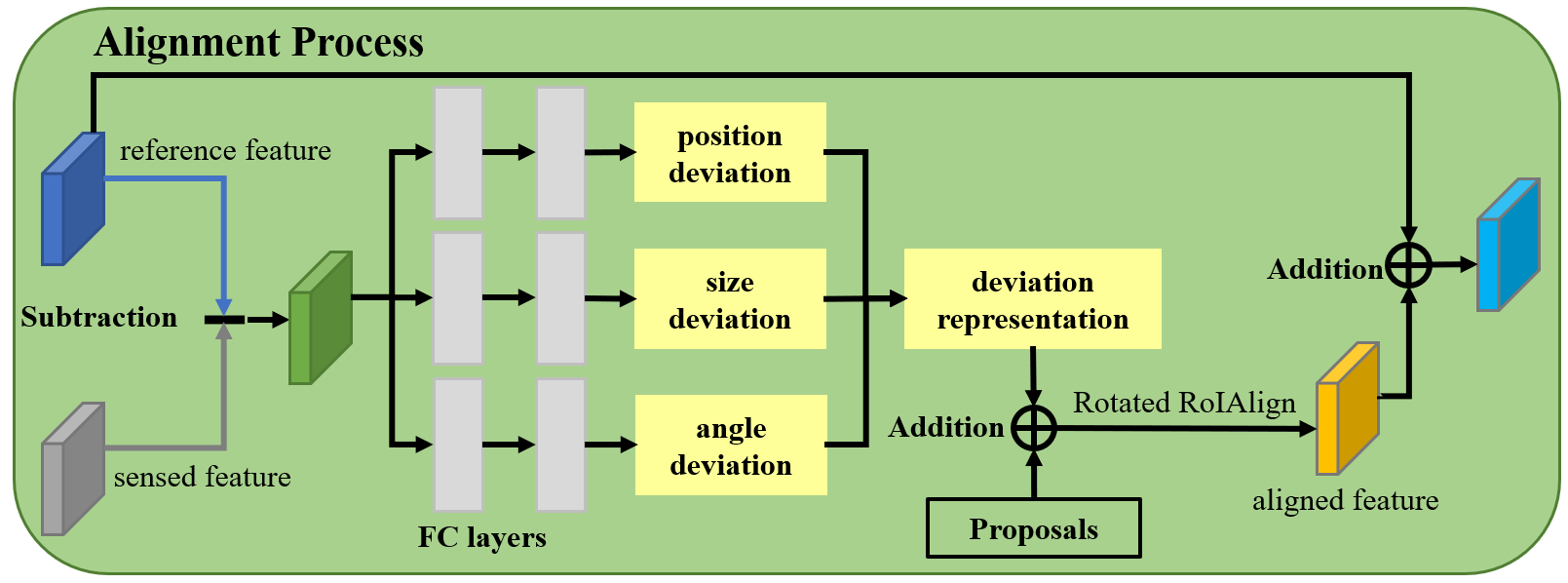}
	\end{center}
    \caption{The concrete structure of the alignment process uses three sets of fully connected layers to predict the deviation of position, size and angle.}
	\label{fig3}
\end{figure*}

Our RGB-IR alignment task is the process of overlaying two proposals of the same object. Refer to  \cite{brown1992survey,zitova2003image}, we introduce the concept of the reference and sensed modality into our task. The alignment process is shown in Fig.~\ref{fig3}. Given the two fixed region feature maps ($\phi_{r}$ and $\phi_{s}$) pooled by rotated RoIAlign operation, we acquire a new feature map $\phi_{d}$ by direct subtraction of two modalities,  $\phi_{d} = \phi_{s}-\phi_{r}$. Through this operation, the feature map $\phi_{d}$ can obtain the differential representation between the two modalities. Then, three sets of consecutive fully connected layers $F_{i}$ are utilized to predict the position deviation $p$, angle deviation $r$ and size deviation $s$ of the region, $\{t(t_{x},t_{y}), s(s_{w}, s_{h}), r(r_{\theta})\}=F_{i}\left(\phi_{d}\right)$. Moreover, we add the predicted deviation to the proposals $p(x, y, w, h, \theta)$ and obtain the sensed proposals $p_{s}(x+t_{x},y+t_{y},w+s_{w},h+s_{h},\theta+r_{\theta})$. Finally, we re-pool the sensed feature maps on the sensed proposals through rotated RoIAlign operation and acquire aligned feature. The final fused feature can be formulated as:
\begin{equation}
\phi_{fused}=\phi_{r}+ROIAlign(p_{s})
\end{equation}

\begin{figure*}[!t]
	\begin{center}
	\includegraphics[scale=0.32]{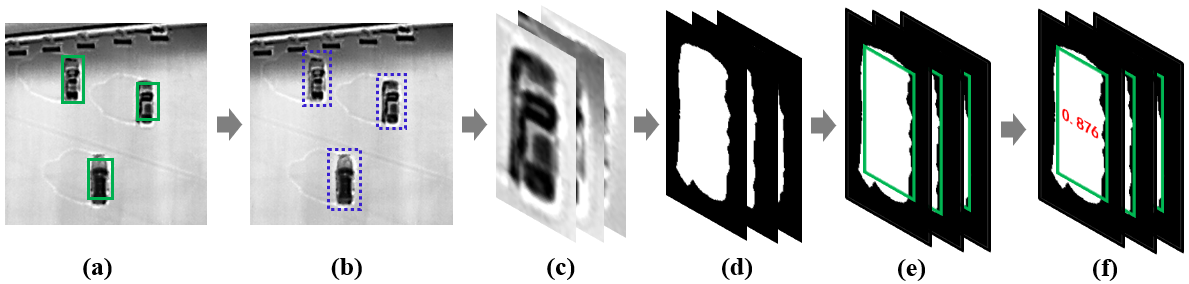}
	\end{center}
    \caption{Illustration of our proposed evaluation method, which can be summarized in the following steps: (a) original bounding-boxes. (b) extend bounding-boxes. (c) crop objects. (d) binarization process. (e) map original bounding-boxes to cropped images. (f) calculate score.}
	\label{fig4}
\end{figure*}

\subsubsection{Modality-Selection Strategy.} 
To alleviate the influence of annotation errors during training, we design the MS strategy to select the bounding-box with better annotation in the two modalities as the reference modality, rather than simply selecting the infrared image as the reference modality  \cite{zhang2019weakly,zhou2020improving,kim2021uncertainty,liu2021cross}. Through this operation, we can determine which bounding-boxes should be used as the reference modality, and then determine the reference feature and sensed feature.

Specifically, we design a evaluation method (shown in Fig.~\ref{fig4}) and perform it on the visible and infrared images separately to select reference bounding-boxes. For each paired bounding-box $B_{rgb}$ and $B_{ir}$, we first extend the bounding-boxes to include the full object. The full objects $C_{rgb}$ and $C_{ir}$ are then cropped from their original images and subjected to color binarization $F_{b}$. Finally, we obtain the scores $S_{rgb}$ and $S_{ir}$ according to their corresponding binary images $F_{b}(C_{rgb})$ and $F_{b}(C_{ir})$, and select the one with the higher score as the reference bounding-box (\emph{e.g.} if $S_{rgb}>S_{ir}$, choose $B_{rgb}$). The score $S$ of the bounding-box is calculated as follows:
\begin{equation}
S=\frac{n}{N_{\text {object }}} \times 0.5+\frac{n}{N_{\text {bounding-box }}} \times 0.5 \label{equ1}
\end{equation}

In equation \eqref{equ1}, $n$ denotes the number of white pixels in the original bounding-box. ${N_{\text {object }}}$ indicates the number of white pixels of the full object. ${N_{\text {bounding-box }}}$ is the total number of pixels of the original bounding-box. $S$ is the final score, and its value range is between 0 and 1. For the ideal bounding-box annotation, $S$ should be close to 1.

\begin{figure*}[!t]
	\begin{center}
	\includegraphics[scale=0.3]{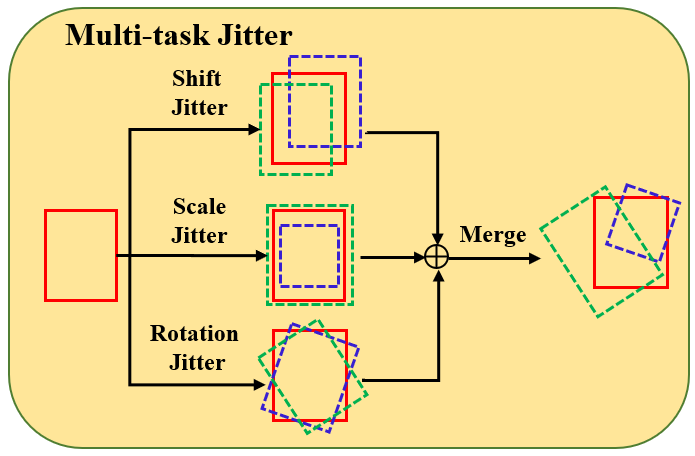}
	\end{center}
    \caption{Illustration of the Multi-task Jitter. Red boxes denote the sensed bounding-boxes. Blue boxes and green boxes represent jitter proposal instances.}
	\label{fig5}
\end{figure*}

\subsubsection{Multi-task Jitter.} 
To improve the robustness of the TSRA module during training, we refer to the ROI jitter strategy \cite{zhang2019weakly} and present a novel Multi-task Jitter (MJ) to augment the deviation. Specifically, the MJ adds the translation, scale and rotation jitter to the sensed proposals, and uses the same settings as \cite{zhang2019weakly} to generate jitter randomly, as shown in Fig.~\ref{fig5}.
% We use a normal distribution to randomly generate the jitter, 
\begin{equation}
\begin{aligned}
&j_{x}, j_{y} \sim N\left(0, \sigma_{x}^{2} ; 0, \sigma_{y}^{2} ; 0\right), \\
&j_{w}, j_{h} \sim N\left(0, \sigma_{w}^{2} ; 0, \sigma_{h}^{2} ; 0\right), \\
&j_{\theta} \sim N\left(0, \sigma_{\theta}^{2}\right),
\end{aligned}
\end{equation}
where $j_{x}$, $j_{y}$, $j_{w}$, $j_{h}$ and $j_{\theta}$ represent location, width, height and angle of a sensed jitter proposal, respectively. 

\subsection{TSRA-based Oriented Detector}
To evaluate our proposed TSRA module, we construct a two-stage oriented object detector incorporating into the TSRA module, called TSFADet. The TSFADet mainly consists of two-stream backbone network, oriented RPN, oriented R-CNN head and our proposed TSRA module. The detailed architecture and description of the loss function are as follows.
\subsubsection{Overall Architecture} \label{overall-architecture}
\begin{figure*}[!t]
	\begin{center}
	\includegraphics[scale=0.26]{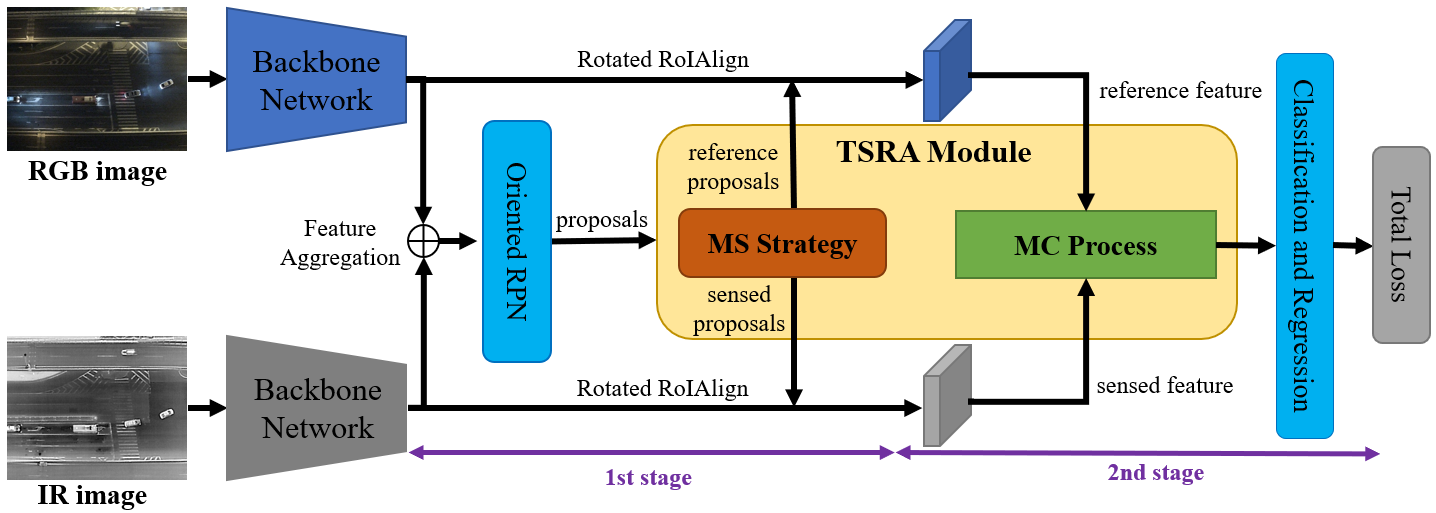}
	\end{center}
    \caption{Overall structure of Two-Stream Feature Alignment Detector (TSFADet), which is a two-stage detector. The first stage generates oriented proposals by oriented RPN and the second stage uses Translation-Scale-Rotation Alignment module to align the oriented features.}
	\label{fig6}
\end{figure*}

The overall architecture of the proposed oriented detector is shown in Fig.~\ref{fig6}. The TSFADet extends the framework of Oriented R-CNN \cite{xie2021oriented} and adopts the two-stream framework to deal with RGB-IR inputs. The backbone is built on FPN follows \cite{lin2017feature}, which produces five levels of features. We aggregate feature maps from two modalities and utilize the Oriented Region Proposal Network (Oriented RPN) to generate proposals. Then we perform MS strategy and alignment process to predict the offset between two modalities. Finally, we acquire the aligned ROI feature to perform the classification and regression task.

\subsubsection{Loss Function.} 
The loss function used for measuring the accuracy of predicted deviation is: 
\begin{equation}
\begin{aligned}
&L_{\text {deviation }}\left(\left\{g_{i}^{*}\right\},\left\{t_{i}\right\},\left\{t_{i}^{*}\right\},\left\{s_{i}\right\},\left\{s_{i}^{*}\right\},\left\{r_{i}\right\},\left\{r_{i}^{*}\right\}\right)= \\
\frac{1}{N_{\text {deviation }}} &\sum_{i=1}^{n} g_{i}^{*}\left(\text { smooth }_{1}\left(t_{i}-t_{i}^{*}\right)+\operatorname{smooth}_{1}\left(s_{i}-s_{i}^{*}\right)+\operatorname{smooth} L_{1}\left(r_{i}-r_{i}^{*}\right)\right)
\end{aligned}
\end{equation}
where $i$ is the index of proposal in a batch, $t_{i}$, $s_{i}$, and $r_{i}$ are the predicted position deviation, size deviation and angle deviation. $g_{i} \in\{0,1\}$, where $g_{i}=1$ if $i$-th proposal is positive, else negative. $N_{deviation}$ is the total number of positive proposals. $t_{i}^{*}$, $s_{i}^{*}$, and $r_{i}^{*}$ are the associated ground-truth position deviation, size deviation and angle deviation of the $i$-th sensed bounding-box, which calculated as follows:

\begin{equation}
\begin{aligned}
\label{equ2}
t_{x}^{*} &=\left(\left(x_{s}-x_{r}\right) \cos \theta_{r}+\left(y_{s}-y_{r}\right) \sin \theta_{r}\right) / w_{r}, \\
t_{y}^{*} &=\left(\left(y_{s}-y_{r}\right) \cos \theta_{r}+\left(x_{s}-x_{r}\right) \sin \theta_{r}\right) / h_{r}, \\
s_{w}^{*} &=\log w_{s} / w_{r}, s_{h}^{*}=\log h_{s} / h_{r}, \\
r_{\theta}^{*} &=\left(\left(\theta_{s}-\theta_{r}\right) \bmod 2 \pi\right) / 2 \pi,
\end{aligned}
\end{equation}
In equation \eqref{equ2}, $\left(x_{s},y_{s},w_{s},h_{s},\theta_{s} \right)$ and $\left(x_{r},y_{r},w_{r},h_{r},\theta_{r} \right)$ are the stack vector for representing location, width, height and angle of the sensed bounding-box and reference bounding-box, respectively.

Finally, the final total loss function can be represented as follows:
\begin{equation}
L=L_{c l s}+L_{r e g}+L_{r p n}+\lambda L_{\text {deviation }}
\end{equation}
where the formulations of $L_{r p n}$, $L_{cls}$ and $L_{reg}$ remain the same as Oriented R-CNN \cite{xie2021oriented}. In our implementation, we set $\lambda=1$, and thus the average gradient of each loss is at the same scale.

\subsubsection{Implementation Details} \label{implemetation}
We implement the network in one unified code library modified from MMDetection  \cite{chen2019mmdetection}. During training, we use the same hyperparameter settings of the original Oriented R-CNN model \cite{xie2021oriented} and use ResNet-50 \cite{he2016deep} as the backbone network, which is pretrained on ImageNet. Horizontal and vertical flipping are adopted as data augmentation during training. The whole network is trained by SGD algorithm with the momentum of 0.9 and the weight decay of 0.0001. We train TSFADet for a maximum of 20 epochs with a batch size of 6 and input image size $512\times512$. The initial learning rate is set to 0.005 and divided by 10 at epoch 16 and 19. The whole framework can be trained end-to-end and the training requires about 14 hours on an NVIDIA GV100 GPU.

\begin{figure*}[!t]
	\begin{center}
	\includegraphics[scale=0.26]{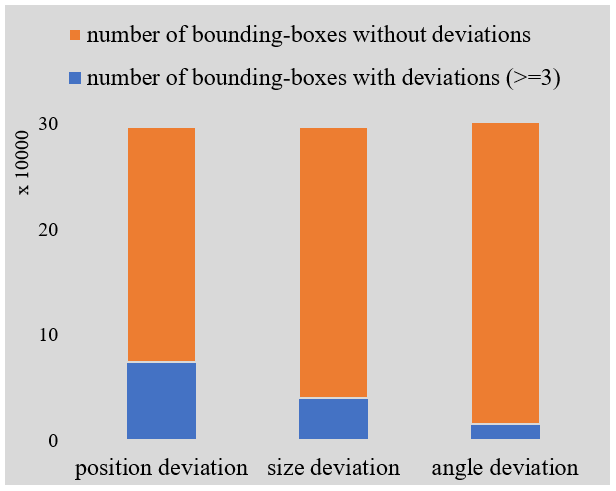}
	\end{center}
    \caption{The statistics of groundtruth bounding-boxes deviation within RGB-IR image pairs in DroneVehicle dataset.}
	\label{fig7}
\end{figure*}

\section{Experimental Results} \label{Experiment}
In this section we show results of experiments we have made to evaluate the effectiveness of TSFADet. In section \ref{dataset}, we first introduce the DroneVehicle dataset \cite{sun2020drone}, and in section \ref{ablation} we carry out ablation studies for the proposed method on the DroneVehicle dataset. In section \ref{camparison} we compare it with other detection approaches.

\subsection{Dataset and Evaluation Metrics} \label{dataset}
Our experiments were conducted on the DroneVehicle RGB-IR vehicle detection dataset \cite{sun2020drone}. DroneVehicle is a large-scale drone-based dataset with well-aligned visible/infrared pairs from day to night. 

The DroneVehicle dataset collects 28,439 RGB-Infrared image pairs, covering urban roads, residential areas, parking lots, and other scenarios. Besides, the authors made rich annotations with oriented bounding boxes for the five categories (car, bus, truck, van and freight car). To verify the effectiveness of our method, we make the following improvements to the annotations of the original dataset:
\begin{itemize}
    \item[$\bullet$] The objects that only annotated in one modality are added in same position to the other modality.
    \item[$\bullet$] If the bounding-box in visible image is under extremely bad illumination, we will discard this bounding-box in both modalities.
    \item[$\bullet$] Sort the bounding-box of the two modalities so that the same object is assigned in the same index.
\end{itemize}
The final training set contains 17,990 image pairs and the validation set contains 1,469 images pairs. We evaluate the detection performance on the validation set and adopt the mean average precision (mAP) as evaluation criteria. For mAP, an Intersection over Union (IoU) threshold of 0.5 is used to calculate True Positives (TP) and False Positives (FP).

\subsection{Statistics of DroneVehicle}
To demonstrate the generality of the weakly misalignment problem in aerial imagery, we also obtain the statistics information of DroneVehicle dataset. we separately count the number of bounding-boxes with deviations (position and size offset by 3 pixels, angle offset by 3 degrees) in the DroneVehicle dataset. As illustrated in Fig.~\ref{fig7}, more than 20\% of the bounding-boxes have the deviation problem. The results show that the weakly misalignment problem in aerial images is a common issue that needs to be considered.

\begin{table*}[!t]
\renewcommand{\arraystretch}{0.9}
\caption{Ablation experiments of TSRA module on DroneVehicle dataset. The symbols of 'P', 'S', and 'A' represent Position, Size and Angle prediction branches respectively. }
\label{table1}
\centering
\begin{tabular}{cccc|cccccc}
\hline
\multicolumn{4}{c|}{}                                                                                                                                  & \textbf{car}         & \textbf{\begin{tabular}[c]{@{}c@{}}freight-car\end{tabular}} & \textbf{truck}       & \textbf{bus}         & \textbf{van}         & \textbf{mAP}   \\ \hline
\multicolumn{4}{c|}{TSFADet (full)}                                                                                                           & \textbf{89.88}       & \textbf{67.87}                                                 & \textbf{63.74}       & \textbf{89.81}       & \textbf{53.99}       & \textbf{73.06} \\ \hline
\multicolumn{4}{c|}{TSFADet w/o MJ}                                                                                                           & 89.34                & 66.53                                                          & 62.65                & 89.62                & 53.67                & 72.36          \\ \hline
\multicolumn{4}{c|}{TSFADet w/o MS}                                                                                                           & 89.78                & 66.26                                                          & 61.44                & 89.60                & 53.17                & 72.05          \\ \hline
\multirow{5}{*}{\begin{tabular}[c]{@{}c@{}}TSFADet \\ w/o MJ and MS\end{tabular}} & P  & S           & A       & \multicolumn{1}{l}{} & \multicolumn{1}{l}{}                                           & \multicolumn{1}{l}{} & \multicolumn{1}{l}{} & \multicolumn{1}{l}{} &                \\ \cdashline{2-4}
                                                                                           & \checkmark          & \checkmark                    & \checkmark                     & 89.69                & 65.11                                                          & 60.39                & 89.43                & 51.01                & 71.13          \\
                                                                                           & \checkmark          & \checkmark                    & \multicolumn{1}{l|}{} & 89.65                & 64.77                                                          & 61.28                & 89.26                & 48.57                & 70.71          \\
                                                                                           & \checkmark          & \multicolumn{1}{l}{} & \checkmark                     & 89.68                & 62.97                                                          & 60.22                & 88.90                & 49.59                & 70.27          \\
                                                                                           & \checkmark          &                      &                       & 89.56                & 62.83                                                          & 58.35                & 89.46                & 47.26                & 69.49          \\ \hline
\multicolumn{4}{c|}{Baseline}                                                                                                                  & 89.45                & 62.14                                                          & 57.00                & 89.09                & 45.43                & 68.62          \\ \hline
\end{tabular}
\end{table*}

\begin{table*}[!t]
\renewcommand{\arraystretch}{0.9}
\caption{Quantitative comparisons of using different methods to demonstrate the contribution of the MS strategy.}
\label{table2}
\centering
\begin{tabular}{c|cccccc}
\hline
\textbf{Methods}         & \textbf{car}   & \textbf{truck} & \textbf{\begin{tabular}[c]{@{}c@{}}freight-car\end{tabular}} & \textbf{bus}   & \textbf{van}   & \textbf{mAP}   \\ \hline
RGB Modality    & 89.47          & 65.56          & 60.36                                                               & 89.63          & 52.82          & 71.57          \\
IR Modality     & 89.78          & 66.26          & 61.44                                                               & 89.60          & 53.17          & 72.05          \\
Random strategy & 89.53          & 66.71          & 62.38                                                               & 89.74          & 53.74          & 72.42          \\
MS strategy     & \textbf{89.88} & \textbf{67.87} & \textbf{63.74}                                                      & \textbf{89.81} & \textbf{53.99} & \textbf{73.06} \\ \hline
\end{tabular}
\end{table*}

\subsection{Ablation Studies}\label{ablation}
Ablation experiments are performed on the DroneVehicle dataset for a detailed analysis in this section. Throughout the experiments, the TSFADet has been inspected by removing each key component from its full version. And the baseline is a two-stream Oriented R-CNN, which only adopts simple additional operation to fuse two modalities. Table~\ref{table1} provides the performance of the different versions with/without the Modality-Selection strategy (MS) and Multi-task Jitter (MJ).

\subsubsection{Alignment Process.}
'TSFADet w/o MJ and MS' version is trained using the alignment process only. As shown in Table~\ref{table1}, 'TSFADet w/o MJ and MS' version obtains 71.13\% mAP, which is almost 3\% higher than the baseline version. Moreover, we also exclude different deviation prediction branches to train the 'TSFADet w/o MJ and MS' version. As can be seen in Table~\ref{table1}, adding Size and Angle prediction branches to the Position prediction branch can improve the performance of the model respectively. Specifically, with the Position and Size prediction branches, the mAP has increased by a significant 2.1\% (from 68.2\% to 70.71\%) compared to the baseline. As a result, the performance improvement of the alignment process was noticeable in DroneVehicle dataset. We conclude that reducing the effect of the weakly misalignment problem caused by hardware errors through the alignment process can improve the multispectral detection performance.

\begin{table*}[!t]
\renewcommand{\arraystretch}{0.9}

\caption{Evaluation results on the DroneVehicle dataset. The last column refers to input modalities of the approach.}
\label{table4}
\centering
% \scriptsize
\begin{tabular}{cccccccc}
\hline
\textbf{Detectors}              & \textbf{car}            & \textbf{truck}          & \textbf{\begin{tabular}[c]{@{}c@{}}freight-car\end{tabular}}    & \textbf{bus}            & \textbf{van}            & \textbf{mAP}            & \textbf{Modality}                \\ \hline
Faster R-CNN(OBB) \cite{ren2015faster}    & 79.69          & 41.99          & 33.99          & 76.94          & 37.68          & 54.06          & \multirow{5}{*}{RGB}    \\
RetinaNet(OBB) \cite{lin2017focal}        & 78.45          & 34.39          & 24.14          & 69.75          & 28.82          & 47.11          &                         \\
ROI Transformer \cite{ding2019learning}   & 61.55          & 55.05          & 42.26          & 85.48          & 44.84          & 61.55          &                         \\
$\rm S^2ANet$ \cite{han2021align}         & 79.86          & 50.02          & 36.21          & 82.77          & 37.52          & 57.28          &                         \\
Oriented R-CNN \cite{xie2021oriented}     & 80.26          & 55.39          & 42.12          & 86.84          & 46.92          & 62.30           &                         \\ \hdashline
Faster-R-CNN(OBB) \cite{ren2015faster}    & 89.68          & 40.95          & 43.10           & 86.32          & 41.21          & 60.27          & \multirow{5}{*}{IR}     \\
RetinaNet(OBB) \cite{lin2017focal}        & 88.81          & 35.43          & 39.47          & 76.45          & 32.12          & 54.45          &                         \\
ROI Transformer \cite{ding2019learning}   & 89.64          & 50.98          & 53.42          & 88.86          & 44.47          & 65.47          &                         \\
$\rm S^2ANet$ \cite{han2021align}         & 89.71          & 51.03          & 50.27          & 88.97          & 44.03          & 64.80           &                         \\
Oriented R-CNN \cite{xie2021oriented}     & 89.63          & 53.92          & 53.86          & 89.15          & 40.95          & 65.50           &                         \\ \hdashline
Halfway Fusion(OBB) \cite{liu2016multispectral}    & 89.85          & 60.34          & 55.51          & 88.97          & 46.28          & 68.19          & \multirow{5}{*}{\begin{tabular}[c]{@{}c@{}}RGB\\ +IR\end{tabular}} \\
CIAN(OBB) \cite{zhang2019cross}             & 89.98          & 62.47          & 60.22          & 88.9           & 49.59          & 70.23          &                         \\
AR-CNN(OBB) \cite{zhang2019weakly}           & \textbf{90.08} & 64.82          & 62.12          & 89.38          & 51.51          & 71.58          &                         \\
TSFADet(Ours) & 89.88          & 67.87 & 63.74 & \textbf{89.81} & 53.99 & 73.06 &                         \\
Cascade-TSFADet (Ours) & 90.01          & \textbf{69.15}  &  \textbf{65.45} & 89.70 & \textbf{55.19} &  \textbf{73.90} &                         \\ \hline
\end{tabular}
\end{table*}

\subsubsection{Modality-Selection Strategy.}
Based on alignment process, we further add the Modality-Selection strategy and validate its contribution. As shown in Table~\ref{table1}, the ’TSFADet w/o MJ’ version is trained with the MS strategy to further improve the detection performance and achieves 72.36\% mAP. To have a deep insight of the effectiveness of Modality-Selection strategy, we investigate the performance of different design choices in Table~\ref{table2}. We use different strategies to select reference modality of bounding-boxes, including directly using the RGB or IR images as the reference modality, randomly selecting bounding-boxes and our MS strategy. From Table~\ref{table2} we can see that our proposed MS strategy has greater advantages. The quantitative comparisons demonstrate the importance of addressing reference bounding-boxes accuracy in the weakly misalignment problem. The experiments show that the MS strategy alleviates the weakly misalignment problem and improves the performance of multispectral detection.

\subsubsection{Multi-task Jitter.}
To verify that the the Multi-task Jitter is effective, we also made a comparison  with/ without it (the 'TSFADet w/o MS' version) in Table~\ref{table1}. It is observed that performance gains can generally be achieved by the Multi-task Jitter. By introducing the Multi-task Jitter which generates various deviations to the sensed bounding-boxes, the alignment process achieves higher accuracy in predicting the deviations between sensed and referenced bounding-boxes. This demonstrates that the detection performance can be further improved by Multi-task Jitter, since it makes the network more robust to solve weakly misalignment problem.

\subsection{Comparisons}\label{camparison}
\subsubsection{Comparison Methods.}
We compare our proposed TSFADet with 5 state-of-the-art single-modality detectors, including Faster R-CNN \cite{ren2015faster}, RetinaNet \cite{lin2017focal}, ROI Transformer \cite{ding2019learning}, $\rm S^2ANet$ \cite{han2021align} and Oriented R-CNN \cite{xie2021oriented}. Since the rotation detectors are focus on detection in single-modality images at present, we re-implement three methods (Halfway Fusion \cite{liu2016multispectral}, CIAN \cite{zhang2019cross} and AR-CNN \cite{zhang2019weakly}) for multispectral object detection on rotation detectors. The backbone of the detectors is also ResNet-50 \cite{he2016deep}. Other hyperparameters including training schedule and data augmentations are also same to the TSFADet to ensure the fairness of the comparisons.

\subsubsection{Quantitative Comparison.}
We evaluate our method and the other eight state-of-the-art methods by using mAP metric. The results are shown in the Table~\ref{table4}. The multispectral methods using both RGB and IR images are superior to the single-modality methods. In single-modality methods, Oriented R-CNN and ROI Transformer both have comparable detection accuracy (65.5\% mAP and 65.47\% mAP) in IR images. We further exploit the advantages of Oriented R-CNN in detection performance and propose TSFADet to perform multispectral detection tasks. From Table~\ref{table4} we can see that our proposed detector achieves 73.06\% mAP, better than other multispectral methods. We also combine Cascade R-CNN \cite{cai2018cascade} structure with our TSFADet and achieve the highest result of 73.90\% mAP.

\begin{table*}[!t]
\renewcommand{\arraystretch}{0.9}
\setlength{\tabcolsep}{3mm}
\caption{Speed versus accuracy on the DroneVehicle dataset.}
\label{table5}
\centering
% \scriptsize
\begin{tabular}{ccccc}
\hline
\textbf{Method}     & \textbf{FPS}  & \textbf{mAP}   & \textbf{Input} & \textbf{framework} \\ \hline
Halfway Fusion(OBB) & 20.4          & 68.19          & RGB+IR         & two-stage         \\
CIAN(OBB)           & \textbf{21.7} & 70.23          & RGB+IR         & one-stage         \\
AR-CNN(OBB)         & 18.2          & 71.58          & RGB+IR         & two-stage         \\
TSFADet(Ours)       & 18.6          & \textbf{73.06} & RGB+IR         & two-stage         \\ \hline
\end{tabular}
\end{table*}

\subsubsection{Speed versus Accuracy}
We compare the speed and accuracy of different detectors on a single NVIDIA GV100 GPU. All detectors are tested with a batch size of 1 under the same settings. During testing, the size of input images is $512\times512$. Table~\ref{table5} reports the comparison results.  Since CIAN is a one-stage detector, its speed is faster than other detectors. In addition, our detector has higher detection accuracy (73.06\% mAP) than other multispectral detectors and runs with comparable speed (18.6 FPS), only 1.8fps slower than the Halfway Fusion method.

\section{Conclusions} \label{Conclusions}
In this work, we propose and analysis the weakly misalignment problem in multispectral aerial detection. Then we explore a TSRA module based multispectral detector named TSFADet to alleviate the weakly misalignment problems. Specifically, we present a new alignment process, which predicts the deviations of position, size and angle to solve the misalignment caused by device factors. Meanwhile, the MS strategy is designed to address the problem caused by human factors. Moreover, we adapt a Multi-task Jitter to further improve the robustness of TSRA module. Our detector can be trained with an end-to-end manner and achieves state-of-the-art accuracy on the DroneVehicle dataset. The proposed method can be generalized to other multispectral detection task and facilitate potential applications.

\section*{Acknowledgments}
This work was supported by National Key R$\&$D Program of China (Grant No.2020AAA0104002) and the Project of the National Natural Science Foundation of China (No.62076018).

\bibliographystyle{splncs04}
\bibliography{egbib}
\end{document}